\documentclass[11pt,a4paper]{article}
\usepackage[hyperref]{acl2021}
\usepackage{times}
\usepackage{latexsym}

\usepackage{microtype}

\usepackage{todonotes}
\usepackage{multirow}
\usepackage{amsmath}
\usepackage{amssymb}
\usepackage{booktabs}
\usepackage{hyperref}
\usepackage[T1]{fontenc}

\aclfinalcopy 

\title{Conditional Generation of Temporally-ordered Event Sequences}

\author{
Shih-Ting Lin$^\spadesuit$ \;\;\;\; Nathanael Chambers$^\diamondsuit$ \;\;\;\; Greg Durrett$^\spadesuit$ \\
 $^\spadesuit$ The University of Texas at Austin\\
 $^\diamondsuit$ United States Naval Academy \\
 {\small \tt j0717lin@cs.utexas.edu, nchamber@usna.edu, gdurrett@cs.utexas.edu}
}

\date{}

\begin{document}
\maketitle
\begin{abstract}
Models of narrative schema knowledge have proven useful for a range of event-related tasks, but they typically do not capture the temporal relationships between events. 
We propose a single model that addresses both temporal ordering, sorting given events into the order they occurred, and event infilling, predicting new events which fit into an existing temporally-ordered sequence. 
We use a BART-based conditional generation model that can capture both temporality and common event co-occurrence, meaning it can be flexibly applied to different tasks in this space.
Our model is trained as a denoising autoencoder: we take temporally-ordered event sequences, shuffle them, delete some events, and then attempt to recover the original event sequence. This task teaches the model to make inferences given incomplete knowledge about the events in an underlying scenario. On the temporal ordering task, we show that our model is able to unscramble event sequences from existing datasets without access to explicitly labeled temporal training data, outperforming both a BERT-based pairwise model and a BERT-based pointer network. On event infilling, human evaluation shows that our model is able to generate events that fit better temporally into the input events when compared to GPT-2 story completion models.
\end{abstract}

\section{Introduction}

\begin{figure}[t]
    \centering
    \includegraphics[width=0.48\textwidth]{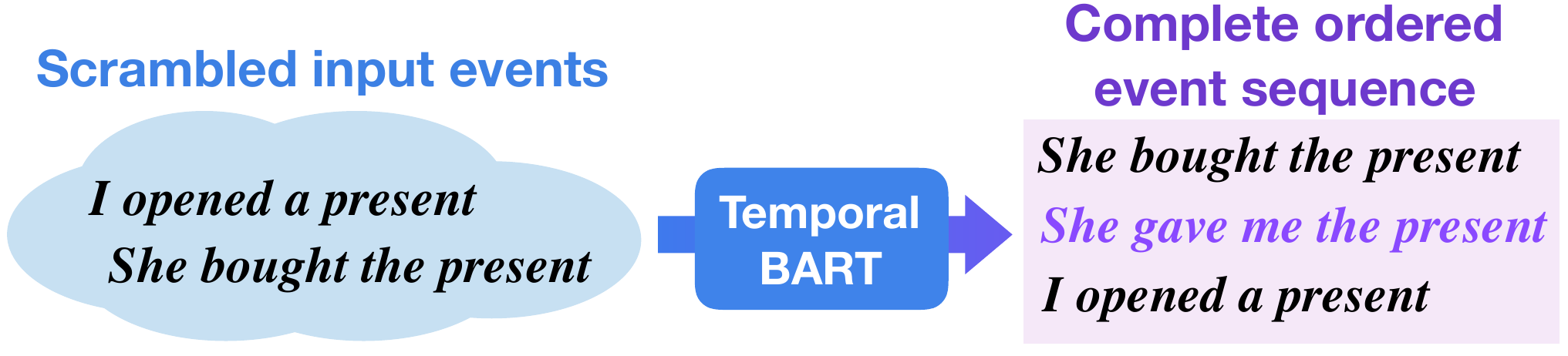}
    \caption{Diagram of our modeling setup: TemporalBART captures both temporal ordering and event cooccurrence to make various event-related inferences.}
    \label{fig:temporalbart-overview}\vspace{-2mm}
\end{figure}

This paper proposes a single model of events to support inferences in two seemingly different tasks: (1) temporal event ordering and (2) event infilling, or inferring unseen or unmentioned events occurring as part of a larger scenario. Figure~\ref{fig:temporalbart-overview} shows an example illustrating these two goals. Unlike prior approaches, we aim to address both with the same model architecture, rather than having to annotate data and build ad-hoc models for each task separately; our goal is to work towards models that capture temporal event knowledge \emph{broadly} and support a wide range of inferences. We thus need a suitably general modeling framework to capture temporal knowledge about events, which in our case will be a BART-based \cite{lewis-etal-2020-bart} model we call TemporalBART. Note that classic temporal relation extraction models, which model temporal ordering \emph{in context} for a particular document, may chiefly learn how to use local discourse cues rather than generalizable event knowledge \citep{chambers-etal-2014-dense, ning-etal-2018-multi}.


The goals in this work relate to past work on learning narrative schemas \citep{Mooney1985LearningSF, chambers-2013-event, peng-roth-2016-two, peng-etal-2017-joint}. Our approach particularly follows a recent line of work using distributed representations of schemas \cite{pichotta2016learning, weber-etal-2018-hierarchical}, which support inferences about events without explicitly materializing a discrete schema library. The target tasks in this work are directly motivated by downstream applications of schema learning. Text generation tasks like story completion rely on understanding what makes narratives plausible and what events might be likely to happen before, after, and between other events \citep{DBLP:journals/corr/JainAMSLS17, DBLP:journals/corr/abs-1811-05701}, motivating our event infilling task. Answering questions about causes, effects, or what might happen next in a scenario requires knowing typical temporal orders of event sequences \citep{zhou-etal-2019-going, Zhou2020TemporalRO, ning-etal-2020-torque}, motivating our temporal ordering task. Prior work has not combined traditional event cooccurrence with event temporality as we do.

We propose a conditional generation model to tackle temporal event ordering and event infilling, and train it as a denoising autoencoder over \emph{out-of-context temporal event sequences}. As shown in Figure~\ref{fig:temporalbart-overview}, the encoder of our TemporalBART model reads a temporally scrambled sequence of a subset of input events, obtained by corrupting a temporally-ordered sequence of events from a corpus. The decoder, which can be viewed as a conditional event language model \citep{kiyomaru-etal-2019-diversity, bosselut-etal-2019-comet, Madaan2020EIGENEI}, then reconstructs the complete, temporally-ordered event sequence. Such denoising training has been successful exploited in many applications \citep{Vincent2010StackedDA, Lu2013SpeechEB, lample2018unsupervised, lewis-etal-2020-bart}, and using seq2seq models to reorder and smooth inputs has been explored before \cite{goyal-durrett-2020-neural}, but to our knowledge we are the first to apply this in this temporal modeling setting. The conditional generation architecture of our model is flexible enough to address a variety of tasks, including our temporal ordering and event infilling tasks, by either sampling from the model or using it to score sequences. Capitalizing on the success of recent pre-trained encoder-decoder transformers \citep{lewis-etal-2020-bart, Raffel2020ExploringTL}, our model itself is based on BART, consuming and producing predicate-argument structures rendered in surface order.

Gathering large-scale high-quality labeled data with temporal annotations is often expensive and requires specially designed annotation schemes \citep{pustejovsky-2003-timeml, cassidy-etal-2014-annotation, ning-etal-2018-multi,ZhaoEtAl2021}. Here, we instead turn to a narrative documents corpus, EventsNarratives \citep{yao-huang-2018-temporal} and design an automatic method to extract the training data we need. In these documents, discourse order is loosely assumed to reflect temporal order, so events extracted from this text can directly provide training data for our models. This use of automatic annotation allows us to use broad-domain data, giving us a strong domain-independent temporal model \cite{ZhaoEtAl2021}.

To evaluate how well our proposed models capture temporal knowledge and solve the two targeted tasks, we apply them on out-of domain test sets in a zero-shot manner. Specifically, for event ordering, we first extract test temporal event sequences from the CaTeRS \cite{mostafazadeh-etal-2016-caters} and MCTaco \cite{zhou-etal-2019-going} datasets, which include the annotations on temporal relations between events. We then compare the performance of our models with two baselines: a BERT-based pairwise model and a BERT-based pointer network. For event infilling, we use the test event sequences from CaTeRS and examine the ability of our models to order unseen events and generate infilled events in comparison with GPT-2 baselines from story generation. Our BART-based models significantly outperform the baseline models on the ordering settings we consider, and human evaluation verifies that our models can generate infilled events that are better temporally-ordered with respect to the input. 

\section{Background and Related Work}

Learning temporal knowledge to order events and generate new events as part of schemas or stories are two problems that have received significant attention, but in contrast to our work, previous work typically focuses on each in isolation.

\subsection{Temporal Event Ordering}
Closely related to the temporal ordering aspect of this paper is temporal relation extraction, which orders pairs of events in text \emph{in document context} \citep{Timebank, cassidy-etal-2014-annotation, ning-etal-2018-multi}. This problem has been addressed as pairwise classification \citep{mani-etal-2006-machine, verhagen-etal-2007-semeval, chambers-etal-2007-classifying, verhagen-pustejovsky-2008-temporal, cheng-miyao-2017-classifying, tourille-etal-2017-neural,goyal-durrett-2019-embedding} 
or as a structured learning problem to enforce constraints on the output \citep{do-etal-2012-joint, ning-etal-2017-structured, ning-etal-2018-improving, leeuwenberg-moens-2017-structured, han-etal-2019-deep, han-etal-2019-joint}. However, even in these latter works, the models focus on pairwise relations. In contrast, our work here views temporal event ordering as a sequence generation problem, which provides models a stronger inductive bias to capture global temporal relations between events. One recent effort \citep{Madaan2020NeuralLM} treats this task as a graph generation problem, and so is able to predict more complex structures, but it focuses solely on ordering and is not suitable for our event infilling goals.

\subsection{Schema Induction}
Schema learning systems are often evaluated on their ability to predict unseen events. Initial work attempted to use statistical methods to derive a library of schematic information \cite{Mooney1985LearningSF,chambers-jurafsky-2008-unsupervised, Jans2012SkipNA}. Another thread exploits event language modeling to learn the distributions over events \citep{pichotta2016learning, peng-roth-2016-two, weber-etal-2018-hierarchical}, or focuses on learning event representations \citep{Modi2016EventEF, Weber2018EventRW} rather than writing down discrete schemas. However, most of this work only models the co-occurrence between events instead of directly considering temporal information, and only represent events as a small tuple of S-V-O headwords.

Another line of work instead directly focuses on extracting coherent narratives from ``story salads'' \citep{wang-etal-2018-picking} or more broadly generating narratives given predefined scenarios \citep{wang-etal-2019-query, qin-etal-2020-back}. However, without considering temporal ordering, these systems are prone to learn discourse ordering of events instead of a strong representation of temporal knowledge.

\begin{figure}[t]
    \centering
    \includegraphics[width=0.34\textwidth]{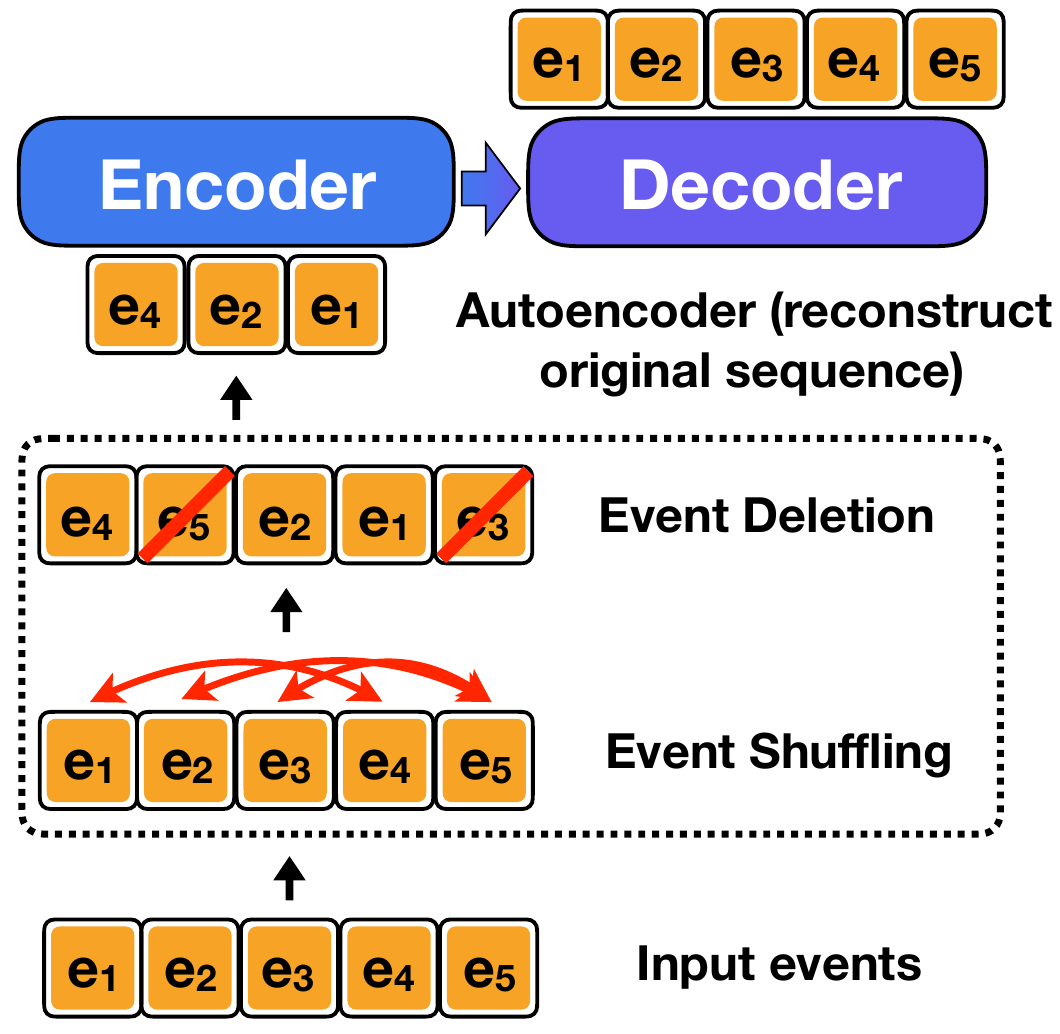}
    \caption{Our event-based denoising autoencoding training scheme used to encourage our model to learn temporal event knowledge. The input is corrupted by shuffling and deletion.}
    \label{fig:training_scheme}\vspace{-2mm}
\end{figure}

\section{Method}

\subsection{Task Formulation and Model}
\label{sec:training_scheme}

Our framework involves modeling a conditional distribution $P(\mathbf{y} \mid \mathbf{x})$ over \textbf{temporal event sequences} $\mathbf{y} = \{e_1,\cdots,e_l\}$, which are sequences of events taken out of context (i.e., not represented as spans in a document) which are part of the same scenario, involve shared actors, and are temporally ordered. The input of the model is a (not necessarily temporal) sequence of events $\mathbf{x} = \{e_1,\cdots,e_m\}$ that represents incomplete information abut the scenario $\mathbf{y}$: a partial set of unordered events. Our model should learn distributions over a true underlying order of events, without obvious gaps in the event sequence, given this incomplete information. By taking events out of context rather than in the context of a document, we are encouraging the model to encode temporal knowledge between events rather than superficial cues like surface textual order or discourse connectives that might determine their order. 

For the definition of events, we follow \citet{chambers-jurafsky-2008-unsupervised} where an event $e$ is a predicate $v_e$ along with its arguments \citep{palmer-etal-2005-proposition}. 

Our model can be formulated as a denoising autoencoder if $\mathbf{x}$ is created as a noised version of $\mathbf{y}$. Specifically, given a temporal event sequence $\mathbf{y}$ as defined above, we first corrupt it to get the required input $\mathbf{x}$ by performing two transformation functions consecutively (see Figure \ref{fig:training_scheme}):

\begin{figure*}[th]
    \centering
    \includegraphics[width=1.0\textwidth]{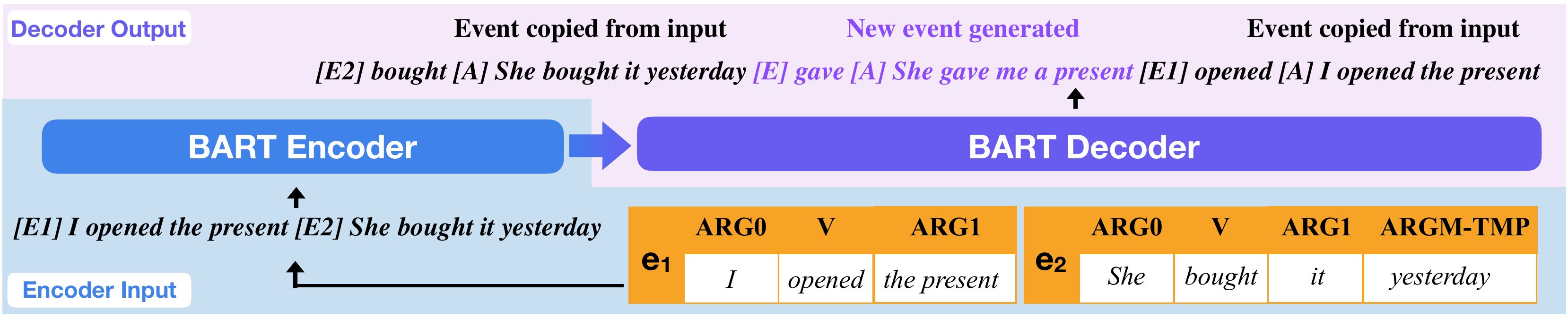}
    \caption{Model architecture of the proposed BART-based conditional generation models. 
    TemporalBART-indexed uses indexed event tags \texttt{[Ei]} as shown in this figure. TemporalBART instead uses the single \texttt{[E]} for all events.}
    \label{fig:model}\vspace{-2mm}
\end{figure*}

\paragraph{Event Shuffling} We first perform a random shuffling of the events in $\mathbf{y}$ to produce $\mathbf{x}$. To perfectly reconstruct the original sequence $\mathbf{y}$, the model must capture the temporal relations between events.

\paragraph{Event Deletion}
We randomly delete each event in $\mathbf{y}$ with probability $p$ to produce $\mathbf{x}$. This denoising scheme is similar to the token deletion transformation in \citet{lewis-etal-2020-bart}. To perfectly reconstruct the original event sequence, the model needs to encode schema-like event knowledge so as to generate events not included in the input $\mathbf{x}$ and insert them at correct positions. As a result, this denoising can help the model learn event infilling.


We train our model to maximize $\log P(\mathbf{y} \mid \mathbf{x})$ on this automatically-constructed data.


\subsection{Model Architecture}
\label{sec:model}
To leverage the power of pretrained transformers, we adopt BART \citep{lewis-etal-2020-bart} as the underlying architecture for our model, and initialize our model with its pretrained weights.

The overall model, shown in Figure \ref{fig:model}, takes a corrupted event sequence $\mathbf{x} = \{e_i\}$ as input, and outputs the true event sequence $\mathbf{y} = \{e_j\}$. To feed the event-based inputs and outputs to BART, we need to represent each event $e$ in a textual format $\mathrm{Repr}(e)$. We represent $e$ with the concatenation of its predicate and all arguments. Unlike previous work which only uses the \emph{syntactic heads} of the predicate and certain arguments \citep{pichotta2016learning, Weber2018EventRW, weber-etal-2018-hierarchical}, our approach preserves complex noun phrase arguments and exposes to the model arguments like temporal modifiers. We strike a balance between using enough information to have meaningful event representations and not consuming entire documents \citep{han-etal-2019-deep, han-etal-2019-joint}, which would result in a model that overly relies on discourse clues.
We then consider two variants for input and output:

\paragraph{TemporalBART}
This model first encodes each event $e_i$ in $\mathbf{x}$ as $\mathrm{Repr}(e_i)$, and concatenates them with a special token \texttt{[E]} prepended in front of each event. This special token can help the model identify the boundary between the input events; such placeholder tokens have been used in related tasks like entity tracking in procedural text \cite{gupta-durrett-2019-tracking}. For the output, we instead prepend \texttt{[E]} $v_{e_j}$ \texttt{[A]} in front of each $\mathrm{Repr}(e_j)$. This setup not only provides an extra supervision signal that encourages the model to predict ordering on the basis of predicates, but also allows us to post-hoc recover an event sequence by checking the predicate part of the generation.

\paragraph{TemporalBART-indexed}
This model, depicted in Figure~\ref{fig:model}, uses the same input and output format as TemporalBART, except the prepended special token \texttt{[E]} is instead \texttt{[Ei]} before each event $e_i$. For the output, if $e_j$ is one of the input events and $e_j = e_i$, then we also change the prepended tokens $e_j$ to \texttt{[Ei]} $v_{e_j}$ \texttt{[A]}. Otherwise, we still use \texttt{[E]} as the special event token. Note that the model is \textit{not} able to ``cheat'' using the \texttt{[Ei]} tokens to do the prediction since the input events are scrambled by the shuffling denoising training scheme described in \S\ref{sec:training_scheme}. Compared to TemporalBART, the use of \texttt{[Ei]} here provides an extra clue for the model to associate input events to output events, which can benefit the event ordering. It also provides a potential way to focus only on modeling the ordering of the target sequence, rather than also mixing in generation decisions, many of which are copying event arguments and often affect the prediction.\footnote{We experiment with this method, which is denoted as ``TemporalBART-indexed (tags only)'', in Appendix \ref{sec:tags only details}} 

Training details of these BART-based models are described in the Appendix.

\subsection{Training Data Collection}
\label{sec:data_collection}
For our framework, the training data we need is event sequences \emph{in temporal order}. Note that most text data occurs in \emph{discourse order}, which is not the same thing: human annotations of temporal relation datasets like TimeBank \citep{Timebank} show that many events mentioned earlier in the text occur later in time. Existing datasets of temporal relations \citep{cassidy-etal-2014-annotation, vashishtha-etal-2019-fine} are small-scale, and annotating more data is expensive and prone to low agreement \cite{ning-etal-2018-multi}. To combat this issue, we instead try to automatically gather the training data we need.

\paragraph{Corpus} We use the  English-language EventsNarratives corpus \citep{yao-huang-2018-temporal}, which contains more than 200,000 \emph{narrative-structured} documents identified from three different source domains including news articles, novel books, and blogs. \citet{yao-huang-2018-temporal} use a weakly supervised method to identify narrative texts, describing a sequence of events in such a way that the discourse order is very likely to reflect the temporal order. This gives us an entry point to collect temporal event sequences automatically from each document. Here we focus on documents in the novel domain as our source for temporal event sequences.

\paragraph{Extracting Temporal Event Sequences} To obtain the training event sequences, we first use an SRL model from AllenNLP \citep{Gardner2017AllenNLP} to extract verbs (events) and their arguments. Then, temporal event sequences are constructed by connecting only the events in different sentences, since the relations between events within the same sentence are unclear even in narrative documents. Here, to ensure all the events in a sequence have a strong relation with each other, we only include chains of events \textbf{that are associated with a common entity} \cite{chambers-jurafsky-2008-unsupervised}, as determined by checking whether the arguments of two event have some shared non-stopword tokens. With this procedure, we are able to collect nearly 2 million temporal event sequences to train on, with nearly 70\% of the sequences consisting of three or more events.

\begin{figure}[t]
    \centering
    \includegraphics[width=0.42\textwidth]{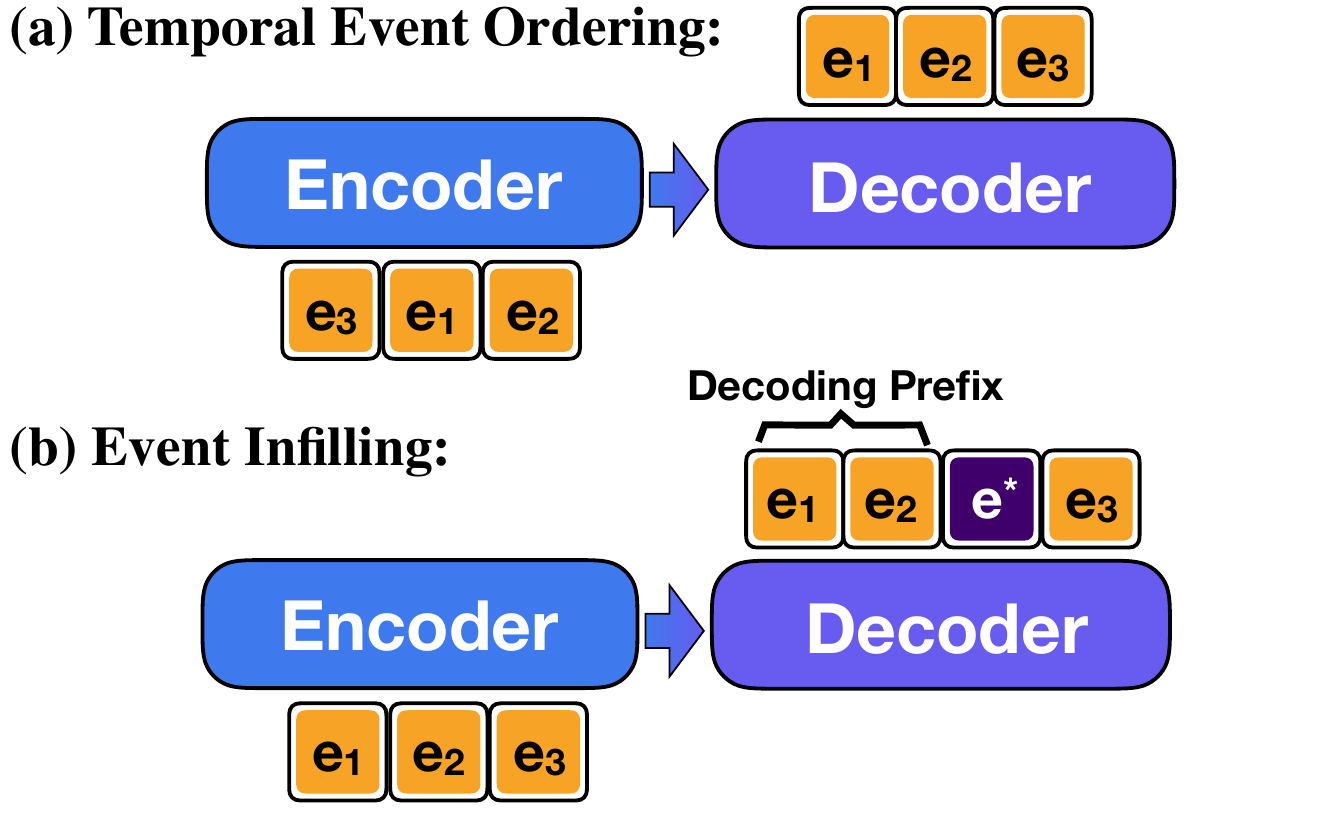}
    \caption{The two targeted tasks in this work: ordering rearranges the set of input events, whereas infilling involves hypothesizing a new event at a specified index.}
    \label{fig:tasks}\vspace{-2mm}
\end{figure}

\section{Target Task Formulation}
\label{sec:target_tasks}
Here we describe the two target tasks of our model and how they can be handled as event-based conditional generation problems. A visual of the task formulations is shown in Figure \ref{fig:tasks}.

\paragraph{Temporal Event Ordering} Given an unordered set of events $\{e_i\}$, this task's goal is to produce the temporal ordering of $\{e_i\}$, as shown in Figure~\ref{fig:tasks}(a). We ask the model to generate an ordered sequence of events $\{e_{f(i)}\}$ given the set $\{e_i\}$, where $f(\cdot)$ is a mapping function to determine the event to put at position $i$. This is a conditional generation problem that is directly solved by our proposed models. 

\paragraph{Event Infilling}
The goal of event infilling is to generate inserted events at some pre-selected insertion positions in a seed event sequence \citep{DBLP:journals/corr/abs-2008-07466}. To simplify the evaluation, here we assume that given an event sequence $\mathbf{x} = \{e_i\}$, models will only be required to generate \emph{one} inserted event at \emph{one} insertion position $i^*$, as shown in Figure~\ref{fig:tasks}(b). We first feed $\{e_i\}$ as the input to our model, then generate one event $e^*$ using $\mathbf{x}_{\mathrm{prefix}} = \{e_i \mid i < i^*\}$ as the decoding prefix. 
To force our models to produce $e^* \notin \mathbf{x}$, we prevent our model from generating $\{v_{e_i}\}$ during the decoding process.

\subsection{Baselines: Temporal Event Ordering}

We compare against two baselines: a state-of-the-art pairwise model used for the in-context temporal ordering task and a pointer network model that directly models event sequence permutations discriminatively.


\paragraph{BERT-based Pairwise Model + SSVM}
We follow the architecture of the Deep SSVM model used in \citet{han-etal-2019-deep} as our first baseline, which tackles event ordering as a pairwise classification problem. This network first exploits a BERT-based model \citep{devlin-etal-2019-bert} to compute pairwise scores for $e_i$ preceding $e_j$ in the output $\mathbf{y}$. The final output is then obtained by 
solving an ILP over all the pairwise scores. 
The overall network is trained with the structured SVM loss so that it can learn to make joint predictions with transitivity constraint. To make this baseline more comparable to our models, we take $\mathrm{Repr}(e_i)$ prepended with \texttt{[E]} as the event representation instead of using the sentence containing $v_{e_i}$ as in \citet{han-etal-2019-deep}.
Detailed formulas are in Appendix \ref{sec:deep ssvm arch}.
We denote this baseline as ``Pairwise+SSVM'' in the evaluations.

\paragraph{BERT-based Pointer Network}
This network first follows the BERT-based Pairwise Model + SSVM to extract the the vectorized representation $\mathbf{U}_{p_{i}}$ for each $e_i$, where $\mathbf{U}$ is the final BERT encoded matrix, and $p_{i}$ is the position of the first token of $e_i$ in the input sequence. These event representations are then instead fed into a LSTM-based pointer network to model the ordering probability by decomposing it in a sequential fashion:
\begin{equation}
    P^{\mathrm{seq}}(\mathbf{y} \mid \mathbf{x}) = \prod_j P( j \mid \mathbf{h}_1, \ldots, \mathbf{U}_{p_{1}}, \ldots)
\end{equation}
$\mathbf{h}_t$ is the decoder hidden state in the pointer network. Compared to the above pairwise baseline, this model has a stronger inductive bias for exploiting global event relations.
We train the sequential model with teacher forcing to maximize the probability of the gold ordering. We denote this baseline as ``BERT-based PN'' in the evaluation section.


\subsection{Baselines: Event Infilling}

\paragraph{HAQAE}
HAQAE \citep{weber-etal-2018-hierarchical} is a vector quantized variational autoencoder which encodes schema knowledge with hierarchical latent variables. Since HAQAE is also an event-level seq2seq autoencoder, we can easily apply it to our setting. During training we follow \citet{weber-etal-2018-hierarchical} except that we use our narrative event sequences for training and represent each event with the predicate-argument format described in \S\ref{sec:model} so it is more comparable to our BART-based models.

\paragraph{GPT-2}
GPT-2 \citep{Radford2019LanguageMA} is a transformer-based pretrained language model  that has been exploited in various generation tasks like story generation \citep{Dathathri2020Plug, rashkin-etal-2020-plotmachines}. However, one issue with the GPT-2 model is that it can only perform uni-directional generation. To apply GPT-2 to generate an inserted event $e^*$, we first concatenate $\{\mathrm{Repr}(e_i) \mid e_i \in \mathbf{x}_{\mathrm{prefix}}\}$ with periods in between, and treat it as the decoding prefix only. We then decode until another period is generated, and take the model's output as the text representation of $e^*$. Except where otherwise specified, we use the GPT2-medium pretrained model from HuggingFace’s Transformer \citep{wolf-etal-2020-transformers}, whose model size is comparable to BART-large.

\paragraph{Infilling GPT-2}
To build a stronger GPT-2 baseline that doesn't only condition on the prefix events, we follow the baselines from \citet{qin-etal-2020-back} to adapt GPT-2 to infilling tasks.  Infilling GPT-2 generates the infilling events by ``wrapping'' the events after the insertion position to the front. That is, the decoding prefix fed to the infilling GPT-2 becomes the concatenation of $\{\mathrm{Repr}(e_i) \mid i >= i^*\}$, \texttt{<SEP>} and $\{\mathrm{Repr}(e_i) \mid i < i^*\}$, again with a period appended after each event. The special token \texttt{<SEP>} is used to help the model to differentiate  the events before and after the insertion position.

\section{Evaluation}

\subsection{Experimental Setup}
All the models used in the evaluation are trained with the temporal event sequences automatically collected on EventsNarratives except GPT-2, since we want to compare the learned knowledge in GPT-2 with our proposed models. Although we are able to gather millions of sequences, for efficiency, we train on 100,000 sequences unless specified otherwise. For each sequence, we extract 2 distinct permutations from the corruption process. This results in 200,000 training examples in total.

During evaluation, all the models are evaluated on out-of-domain datasets in a zero-shot way, i.e., no fine-tuning is performed on the evaluation sets. 

\subsection{Temporal Event Ordering}
\label{sec:eval event ordering}

\begin{figure}[t]
    \centering
    \includegraphics[width=0.48\textwidth]{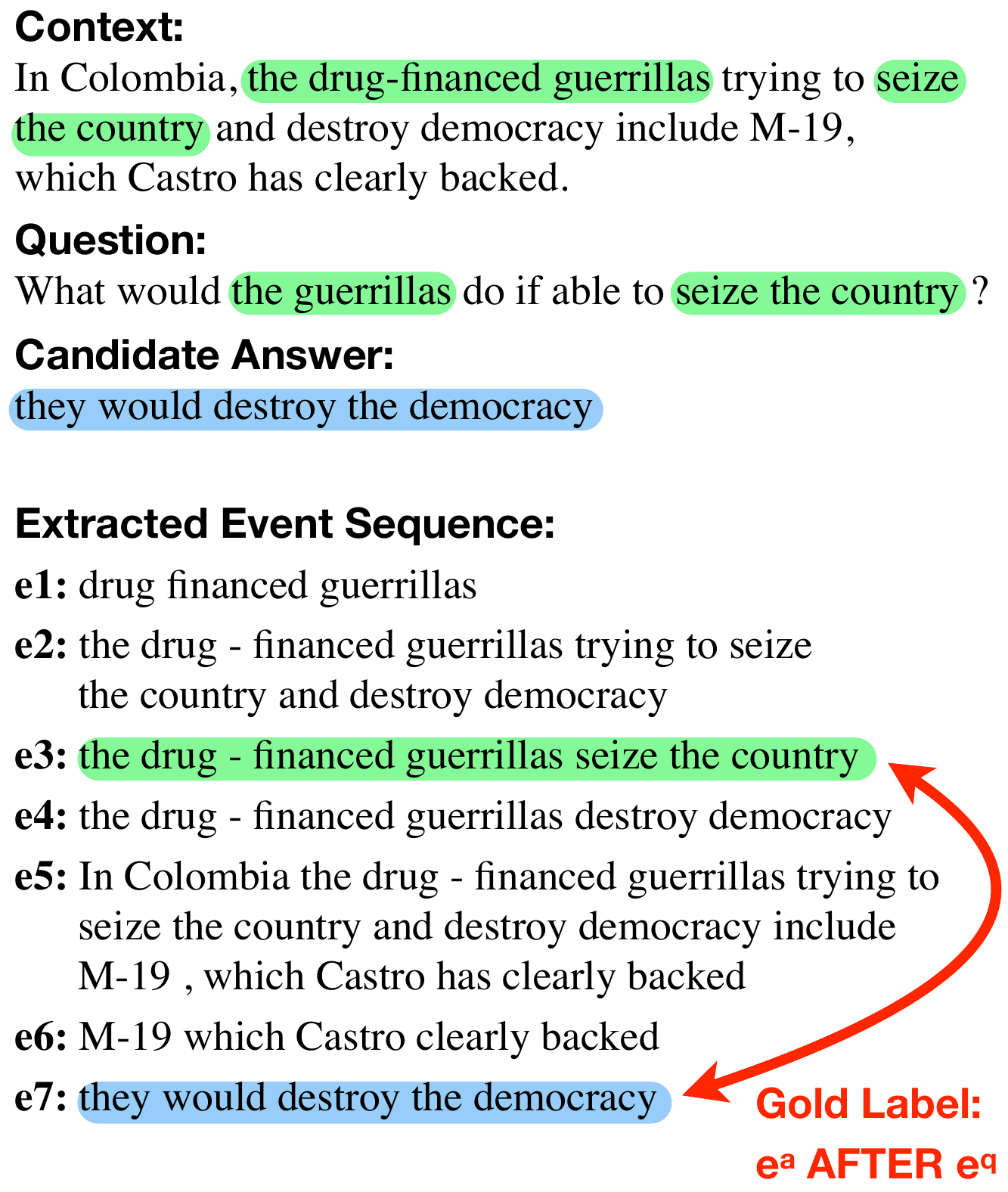}
    \caption{An example of the event sequence extracted from a context-question-answer tuple in MCTaco. $e^q$ and $e^a$ are highlighted with the color green and blue respectively.}
    \label{fig:mctao_example}
\end{figure}

\subsubsection{Datasets}
\label{sec:eval datasets}
We use two out-of-domain English datasets to extract the test temporal event sequences: CaTeRS and MCTaco. As during training, two different permutations are produced from each extracted sequence.

\paragraph{CaTeRS \citep{mostafazadeh-etal-2016-caters}} CaTeRS includes annotations of events and their casual and temporal relations on 320 five-sentence short stories sampled from the ROCStories corpus \citep{mostafazadeh-etal-2016-corpus}. 
To extract the evaluation data from CaTeRS, we first apply the SRL model used in \S\ref{sec:data_collection} on each story. Then, a directed acyclic graph is constructed with a node being an event $e$ whose predicate $v_e$ can be captured by the SRL model, and an edge $(e_i, e_j)$ indicating $e_i$ happens temporally before $e_j$. Note that here we treat all types of annotated relations except ``IDENTITY'', ``DURING'' and ``CAUSE\_TO\_END'' as ``BEFORE'', as suggested in \citet{mostafazadeh-etal-2016-caters}. Test temporal event sequences are then extracted by retrieving all the path from the source nodes to sink nodes in the graph.
With this procedure, we are able to gather 842 event sequences, 60\% of which contain 3 or more events. With permutations, the final CaTeRS evaluation set has 1684 examples.

\paragraph{MCTaco \citep{zhou-etal-2019-going}} MCTaco is a multiple-choice QA dataset for evaluating model understanding on 5 different types of temporal commonsense. To extract suitable test data, we focus on questions with the reasoning type of ``event ordering'' and their positive candidates. Each data point here consists of a sentence describing multiple events $\{e^c_i\}$, a question asking what event could happen temporally before/after a particular event $e^q \in \{e^c_i\}$, and a candidate event $e^a$. Critically, \emph{the question itself} tells us whether $e^a$ should happen before/after $e^q$ in the temporal event sequence formed by $\{e^c_i\} \cup \{e^a\}$.

With this annotation, we evaluate our models by first feeding the randomly shuffled $\{e^c_i\} \cup \{e^a\}$ into a model, then checking the ordering between $e^a$ and $e^q$ in the output sequence. Here, we were able to extract 585 test sequences from MCTaco. For each sequence, $\{e^c_i\}$ and $e^a$ are extracted with the SRL model used in \S\ref{sec:data_collection}. For the question, we first use a set of pre-defined regex templates to extract an event $e^q$ and a temporal relation (``before'' / ``after''). We then match $e^q$ to one of $e^c_i$ by ROUGE-L scores. See Figure \ref{fig:mctao_example} for an example of the extracted data.

Compared to CaTeRS, since the sentences here are from 9 different domains in MultiRC \citep{khashabi-etal-2018-looking}, the types of events are more diverse. The event arguments are also more complex.


\begin{table}[t]
    \centering
    \small
    \renewcommand{\tabcolsep}{0.8mm}
    \begin{tabular}{l|ccc|ccc}
      \toprule
      \multirow{2}{*}{Architecture} & All & Length >= 3 \\
      & Pairwise Acc. & Pairwise Acc. \\
      \midrule
      Random & 50.4 & 50.2 \\
      Pairwise+SSVM & 65.7 & 62.3 \\
      BERT-based PN & 54.1 & 52.3 \\
      \midrule
      TemporalBART & 77.1 & 74.7 \\
      TemporalBART-indexed & \textbf{79.7} & \textbf{78.0} \\
      \bottomrule
    \end{tabular}
    \caption{Averaged pairwise accuracy between the gold and predicted ordering generated by each model on temporal event sequences from CaTeRS. The rightmost column is sequences with 3+ events. 
    }
    \label{tab:caters_ordering}\vspace{-2mm}
\end{table}

\subsubsection{Results on CaTeRS}
We first examine the temporal ordering results on CaTeRS, shown in Table  \ref{tab:caters_ordering}. We compute the pairwise accuracy of the predicted event sequences, or how many pairs of events in the output are ordered correctly by a model. Note that the BART-based models can deviate from generating permutations of the input; however, we found that the most probable generated sequences were almost exact permutations of the input or easily aligned to the input using a heuristic.

Our BART-based models outperform the BERT-based pointer network by more than 20 points, a huge margin. One possible reason is that the decoder of BART can condition on the token-level embeddings of the events when generating the output events, whereas in the pointer network, the decoder is only aware of the condensed event embeddings $\mathbf{U}_{p_{i}}$. Our two BART-based models also outperform the BERT-based pairwise model on both all sequences and long sequences. 

\begin{table}[t]
    \centering
    \small
    \begin{tabular}{l|cc}
      \toprule
      Architecture & Acc. & Macro F1 \\
      \midrule
      Majority & 90.6 & 47.5 \\
      Pairwise+SSVM & 67.2 & 47.0 \\
      BERT-based PN & 54.7 & 42.7 \\
      \midrule
      TemporalBART & 63.9 & 50.1 \\
      TemporalBART-indexed & \textbf{74.9} & \textbf{55.1} \\
      \bottomrule
    \end{tabular}
    \caption{Temporal ordering results on MCTaco sequences. Metrics are computed on the ordering between the answer event and sentence event. The test set is imbalanced, so we include macro F1. Our BART-based models outperform the baselines for macro F1.}
    \label{tab:mctaco_ordering}\vspace{-3mm}
\end{table}

\subsubsection{Results on MCTaco}
Results on MCTaco are shown in Table \ref{tab:mctaco_ordering}. Here since we only know the gold temporal relation of one pair of events in the input, i.e $e^q$ and $e^a$, the averaged accuracy on predicting the order of $e^q$ and $e^a$ is computed. In addition, since the ratio of before/after questions is significantly unbalanced in MCTaco, with 90\% asking about the ``after'' relationship, we also compute the macro F1 score as our metric (averaging F1 across these two classes).

Our two baselines perform worse than just picking the majority label. This is possibly due to the high diversity of events in MCTaco, which makes it much harder to apply a zero-shot model. In contrast, TemporalBART achieves an F1 score about 3 points higher than the Pairwise+SSVM baseline, and TemporalBART-indexed further performs best among all.

In Appendix \ref{sec:timex}, we also show that our models are able to learn temporal phenomenon not explicitly annotated in our training data, which is another demonstration of our model's ability to generalize.

\subsection{Ordering Unseen Events}
\label{sec:eval_unseen_events}

We evaluate our BART-based models on an additional variant of this ordering problem that better tests their capability as \emph{generative} models. Recall that previously, BART conditions on the complete (but possibly scrambled) sequence of events.
We now consider ordering an event in the decoder that the model \emph{does not} condition on in the encoder. Concretely, for each temporal event sequence in CaTeRS, we randomly select one event $e^*$, and treat the rest of the sequence as the seed input event sequence $\{e_1, \cdots, e_N\}$. Then we check if a model can correctly determine where to insert $e^*$ into the input sequence. Specifically, for both the BART-based models and the GPT-2 baselines, we use the generation probability to rank event sequences $\{e_1, \cdots, e_{i^*-1}, e^*, e_{i^*}, \cdots, e_N\}$ for $i^*$ between 1 and $N+1$ (all possible locations). 
If a model correctly ranks the gold candidate higher, it indicates that it can model temporal relations between seen events and new \emph{unseen events} it may generate.

The results are shown in Table \ref{tab:caters_order_extra_one}, where we compute the top-1 and top-2 exact match (EM): did the model rank the gold sequence 1st or 2nd highest? Our GPT-2 variants are only slightly better than random. HAQAE, also using an autoencoder framework, performs worse than infilling GPT-2, likely due to the lack of large-scale pretraining and the loss of information when compressing input into latent variables. Our BART-based models are significantly better, with TemporalBART-indexed showing the benefit of using indexed event markers to help the model capture order. 
We also perform an ablation of deletion during training (Figure~\ref{fig:training_scheme}). Unsurprisingly for this unseen event evaluation, not deleting events in training (setting $p$ to 0) causes a major drop by 14 EM points. Deletion denoising is evidently critical to model new events. 

\begin{table}[t]
    \centering
    \small
    \renewcommand{\tabcolsep}{0.8mm}
    \begin{tabular}{l|cc|cc}
      \toprule
      \multirow{2}{*}{Architecture} & \multicolumn{2}{c|}{All} & \multicolumn{2}{c}{Length >= 3} \\
      & EM & Top2 EM & EM & Top2 EM \\
      \midrule
      Random & 34.1 & 69.5 & 23.7 & 48.7 \\
      HAQAE & 37.1 & 71.9 & 28.7 & 53.2 \\
      GPT-2 & 35.2 & 68.4 & 22.6 & 48.2 \\
      Infilling GPT-2 & 38.8 & 73.5 & 26.3 & 55.4 \\
      \midrule
      TemporalBART & 57.7 & 83.3 & 48.2 & 70.6 \\
      TemporalBART-indexed & \textbf{58.4} & \textbf{87.4} & \textbf{50.9} & \textbf{77.4}
      \\
      - event deletion & 42.4 & 73.0 & 29.8 & 53.8 \\
      \bottomrule
    \end{tabular}
    \caption{Comparison of the ability to tackle unseen events between our BART-based models and baselines on CaTeRS. The right columns are computed on test sequences of 3 or more events.
    }
    \label{tab:caters_order_extra_one}\vspace{-3mm}
\end{table}

\begin{figure*}[t]
    \centering
    \includegraphics[scale=0.47]{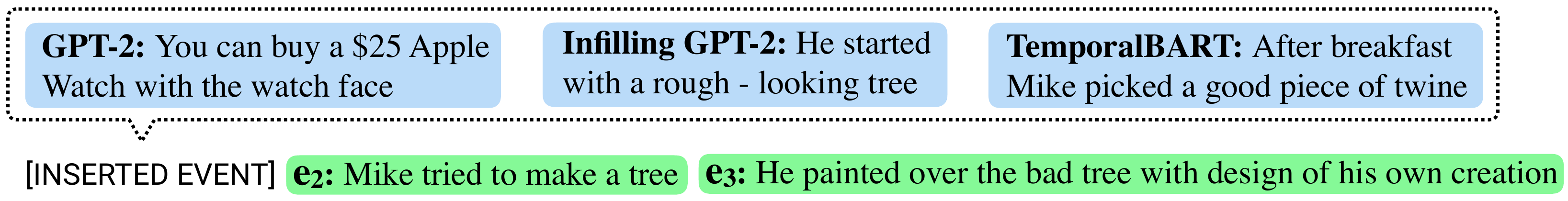}
    \caption{Real examples of infilled events generated by GPT-2, infilling GPT-2 and TemporalBART respectively. Green events are the input, and blue events are the infilled events generated by the models.}
    \label{fig:infilling_examples}\vspace{-2mm}
\end{figure*}

\subsection{Event Infilling}
\label{sec:event_infilling}


Now we turn to temporal event infilling: given a CaTeRS sequence, remove a random event at index $i^*$, and denote the resulting sequence $\{e_1, \cdots, e_N\}$. We then ask a model to generate one event $e^*$ at position $i^*$ so $\{e_1, \cdots, e_{i^*-1}, e^*, e_{i^*}, \cdots, e_N\}$ is temporally ordered with the new event. 

We evaluate the quality of the generated (inserted) events by human evaluation on Amazon Mechanical Turk. Specifically, we randomly sample 30 examples from CaTeRS and have 5 raters judge the coherence and temporality (on a scale from 0 to 2) of the inserted event from each model. 
See Figure \ref{fig:mturk_prompt} in Appendix for our exact prompt. The final scores for each model on coherence and temporality are computed by taking the average of the majority rating on each prediction. Here we only include GPT-2 models as baselines since HAQAE is also using the autoencoder framework, and already performs significantly worse in \S\ref{sec:eval_unseen_events}.  

\begin{table}[t]
    \centering
    \small
    \begin{tabular}{l|cc}
      \toprule
      Architecture & Coherence & Temporality \\
      \midrule
      GPT-2 & 1.37 & 0.57 \\
      Infilling GPT-2 & \textbf{1.50} & 0.87 \\
      \midrule
      TemporalBART & 1.43 & \textbf{1.10} \\
      TemporalBART-indexed & \textbf{1.50} & 1.03 \\
      \bottomrule
    \end{tabular}
    \caption{Human evaluation of event infilling (0-2 scale). Data are event sequences from CaTeRS. All models fill in coherent events, but our BART-based output is more temporally ordered with respect to the input events.}
    \label{tab:caters_human_eval}\vspace{-3mm}
\end{table}


The results of this evaluation are shown in Table \ref{tab:caters_human_eval}. All models achieve reasonable coherence scores.
However in terms of temporality, GPT-2 performs worst, as expected, since it can only condition on partial input event sequences while the other three consider the whole event sequence as input. Both of the BART-based models achieve better performance than infilling GPT-2. The improvements on the temporal score are significant with $p<0.05$ according to bootstrap resampling for both TemporalBART models with respect to infilling GPT-2. 

Figure \ref{fig:infilling_examples} gives examples of infilled events generated by GPT-2, infilling GPT-2, and TemporalBART. On this specific test example, GPT-2 generates an event generally about the Apple watch, which is less relevant to the input scenario about Mike making a tree. The event generated by infilling GPT-2 is coherent with the scenario, but doesn't occur in the correct order with respect to the input events. The event generated by TemporalBART is the best in terms of coherence and temporality. 
More examples are in Table \ref{tab:more_infilling_examples} of the Appendix.

\begin{figure}[t]
    \centering
    \includegraphics[width=0.48\textwidth]{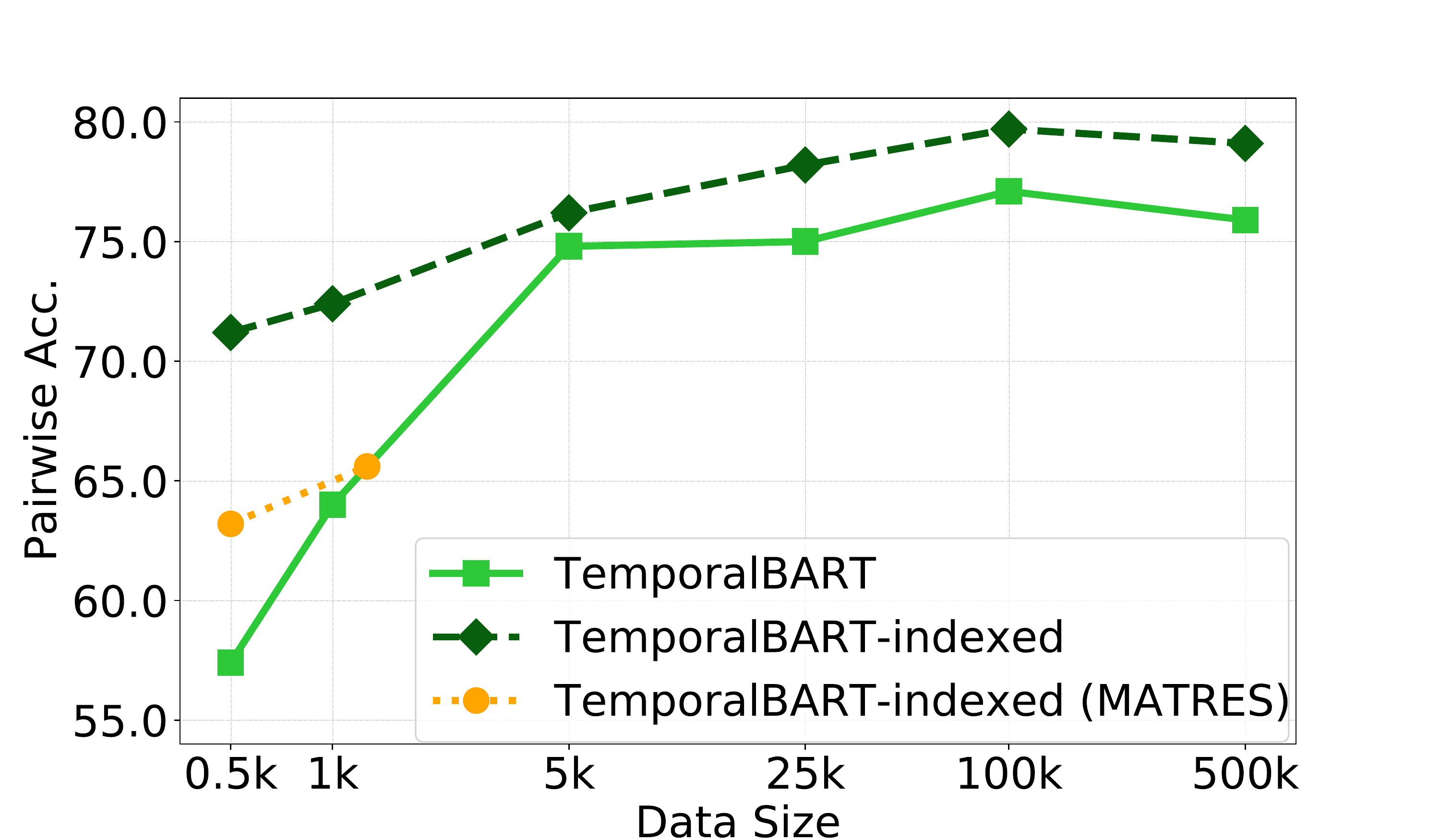}
    \caption{Pairwise accuracy of the two proposed BART-based models on temporal event ordering on CaTeRS when trained with different numbers of event sequences from narrative documents and MATRES.
    }
    \label{fig:data_size}\vspace{-3mm}
\end{figure}

\subsection{The Effectiveness of Narrative Data}

Figure \ref{fig:data_size} shows that the performance of both our models on the CaTeRS ordering task improves when increasing the amount of narrative training data. This demonstrates that the automatically extracted temporal event sequences are useful and diverse enough to help the models to learn temporal-related knowledge. The TemporalBART-indexed model is effective on surprisingly small amounts of data, but also scales well with data size; however, we observe a plateau in both models which motivated our decision to use 100k training sequences. 

For comparison, we train our TemporalBART-indexed model on 1266 event sequences gathered from the MATRES dataset, a human-labeled dataset for temporal relation extraction, using the same procedure we applied to CaTeRS. However, Figure \ref{fig:data_size} shows that the resulting performance, 65.6 on MATRES, is significantly lower than the best number we get on narrative data. Even with the same size training set, using narrative data achieves over 7 points of improvement over using MATRES. This suggests that the small-scale human-labeled dataset is not enough for models to learn \emph{generalized} temporal knowledge, and even with the same amount of data, narrative data may be a better source for general temporal knowledge.

\section{Conclusion}
This work presents a BART-based conditional generation model and a denoising autoencoder framework to learn temporal event knowledge, and addresses \emph{both} temporal ordering and event infilling tasks by pretraining on automatically collected data. Our experiments demonstrate that our model is able to perform temporal ordering and infilling in a zero-shot manner, not fine-tuned on our target datasets, which suggests that it can also be applied to other settings requiring event schematic and temporal knowledge.

\section*{Acknowledgments}

Thanks to Mahnaz Koupaee from Stony Brook University for providing directions on our HAQAE baseline and to the members of the UT TAUR lab for helpful discussion, particularly Yasumasa Onoe and Jiacheng Xu for suggestions on the human evaluation. Thanks as well to the anonymous reviewers for their comments. This work is based on research that is in part supported by the Air Force Research Laboratory (AFRL), DARPA, for the KAIROS program under agreement number FA8750-19-2-1003. The U.S. Government is authorized to reproduce and distribute reprints for Governmental purposes notwithstanding any copyright notation thereon. The views and conclusions contained herein are those of the authors and should not be interpreted as necessarily representing the official policies or endorsements, either expressed or implied, of the Air Force Research Laboratory (AFRL), DARPA, or the U.S. Government.

\bibliographystyle{acl_natbib}
\bibliography{anthology,acl2021}

\clearpage
\appendix

\section{Scoring Orderings with TemporalBART-indexed (tags only)}
\label{sec:tags only details}
TemporalBART-indexed (tags only) scores whether an output sequence $\mathbf{y}$ is temporally ordered by gathering the generation scores on the special tokens \texttt{[Ei]} only as its final ordering score:
\begin{equation}
    P^{\mathrm{tag}}(\mathbf{y} | \mathbf{x}) = \prod_{t \in \mathbf{I}} \mathrm{BART}(w^{\mathbf{y}}_{t} | \mathbf{x}, w^{\mathbf{y}}_{1}, \cdots, w^{\mathbf{y}}_{t-1})
\end{equation}
where $\{w^\mathbf{y}_t\}$ is the text representation of $\mathbf{y}$ and $\mathbf{I}$ is the set of the positions of the special tokens \texttt{[Ei]} in $\{w^\mathbf{y}_t\}$. This allows us to make a judgment only depending on the predicted \emph{temporal order} of the events rather than mixing in general token order. In contrast, TemporalBART scores a sequence of events $\mathbf{y}$ with the generation probability on the entire text representation of $\mathbf{y}$:
\begin{equation}
    P^{\mathrm{gen}}(\mathbf{y} | \mathbf{x}) = \prod_t \mathrm{BART}(w^{\mathbf{y}}_{t} | \mathbf{x}, w^{\mathbf{y}}_{1}, \cdots, w^{\mathbf{y}}_{t-1})
\end{equation}
Since many of the generation decisions here are copying event arguments, the prediction could be largely affected by the correlation of tokens within each argument.

\begin{table}[h]
    \centering
    \small
    \renewcommand{\tabcolsep}{1.0mm}
    \begin{tabular}{l|cc}
      \toprule
      Architecture & Acc. & Macro F1 \\
      \midrule
      TemporalBART-indexed & 74.9 & 55.1 \\
      TemporalBART-indexed (tags only) & \textbf{76.6} & \textbf{56.4} \\
      \bottomrule
    \end{tabular}
    \caption{The comparison between TemporalBART-indexed and its (tags only) variant on temporal event ordering. The test data and metrics used here are same as in Table \ref{tab:mctaco_ordering}.}
    \label{tab:mctaco_ordering_tag_only}
\end{table}

We evaluate ``TemporalBART-indexed (tags only)'' on the temporal event ordering with the procedure used for the models in Table \ref{tab:mctaco_ordering}. Table \ref{tab:mctaco_ordering_tag_only} shows that this (tags only) variant further boosts the performance of TemporalBART-indexed by 1.3 points on the macro F1. This result verifies that this setting can help prevent the ordering scores from being overly affected by the text generation probabilities, which is particularly important for MCTaco, where the arguments of events are more complex. 

\section{Architecture of BERT-based Pairwise Model + SSVM}
\label{sec:deep ssvm arch}
This network uses a BERT-based model \citep{devlin-etal-2019-bert} to obtain a vectorized representation for each input event $e_i$ in $\mathbf{x}$. As with the BART-based models, the input to the BERT model is the concatenation of $\mathrm{Repr}(e_i)$ with \texttt{[E]} being prepended in front of each event. The vectorized representation for $e_i$ is then extracted by $\mathbf{U}_{p_{i}}$, where $\mathbf{U}$ is the final BERT encoded matrix, and $p_{i}$ is the position of the first token of $e_i$ in the input sequence. Each pair of event representations, $\mathbf{U}_{p_{i}}$ and $\mathbf{U}_{p_{j}}$ are then fed into a feed-forward function $g$ to compute a score $B$ for $e_i$ preceding $e_j$ in the output $\mathbf{y}$:
\begin{equation}
    B(e_i, e_j) = g([\mathbf{U}_{p_{i}}; \mathbf{U}_{p_{j}}; \mathbf{U}_{p_{i}}\odot\mathbf{U}_{p_{j}}])
\end{equation}
Finally, the final output $\mathbf{y}$ is computed by finding the best permutation over all of the pairwise scores by solving an ILP.

\section{Training Details of BART-based Models}
We train our BART-based conditional generation models to minimize negative log likelihood of reconstructing the original event sequence. We set the learning rate to 1e-5, and use a polynomial decay scheduling with 500 steps of warm-up. All of the models are trained for 10 epochs, with each epoch being 2000 updates and the batch size being 64. For the deletion training scheme, we set the event deletion probability $p$ to 0.15. The framework is implemented with PyTorch \citep{NEURIPS2019_9015} and AllenNLP \citep{Gardner2017AllenNLP}, and we use the BART-large pretrained model from  HuggingFace’s
Transformers library \citep{wolf-etal-2020-transformers}.

During the evaluation for temporal event ordering, we decode the output event sequences using beam search with the beam size being 4. For event infilling task, we use nucleus sampling with $p$ set to 0.8.

\begin{figure}[t]
    \centering
    \includegraphics[width=0.43\textwidth]{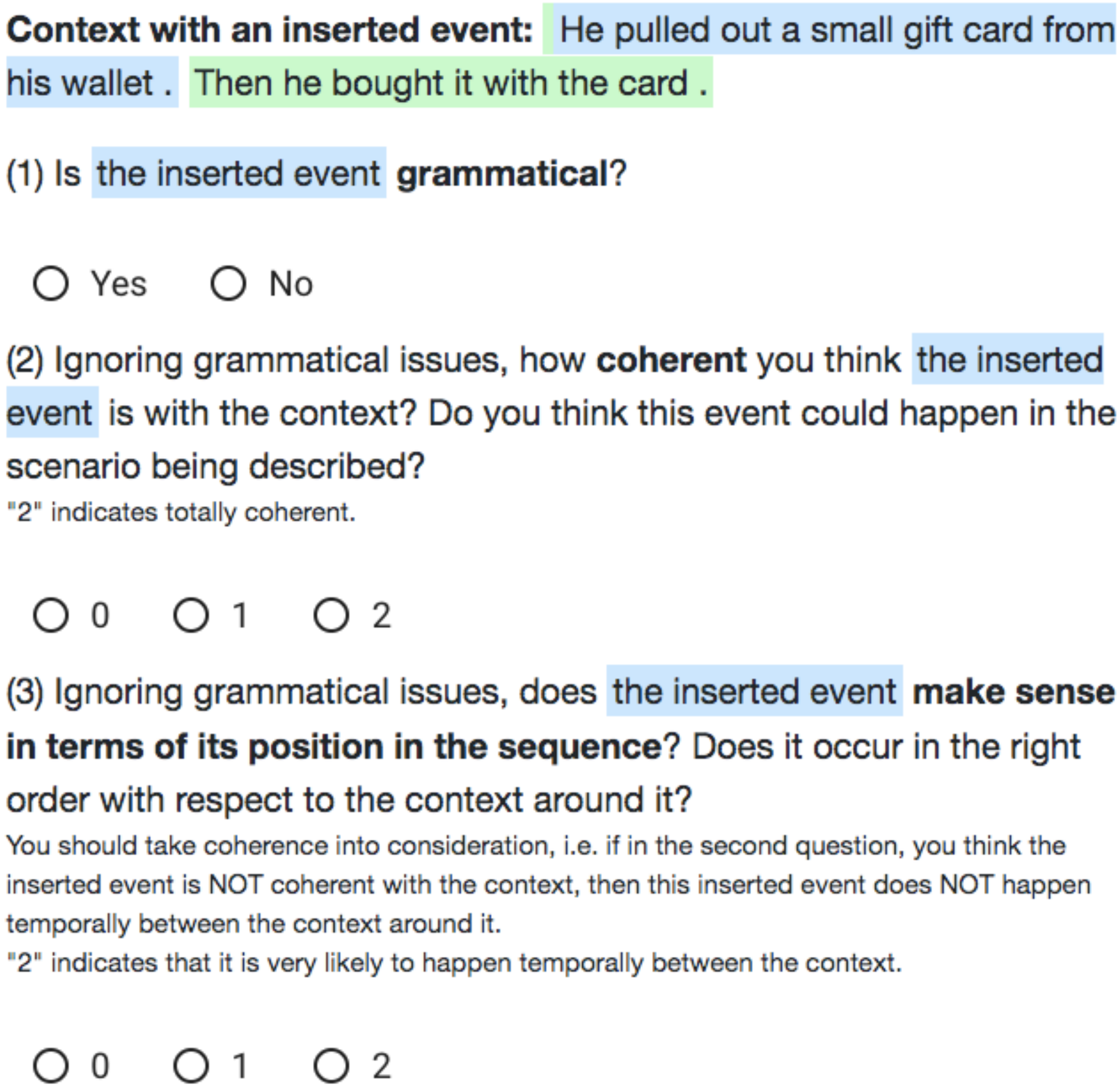}
    \caption{A screenshot of prompt for the human evaluation described in \S\ref{sec:event_infilling}, which includes the 3 questions the raters are asked to judge the event infilling outputs from each model. The input events are highlighted with the color green, and blue for the inserted events.}
    \label{fig:mturk_prompt}
\end{figure}
\section{Human Evaluation}
Figure \ref{fig:mturk_prompt} shows the prompt for the human evaluation described in \S\ref{sec:event_infilling}, where we ask the MTurk raters to evaluate the coherence and temporality of the generation outputs. To help the raters ignore grammatical issues when making decisions, we first ask them to check the grammaticality, then separately judge the coherence and the temporality.

\begin{figure}[t]
    \centering
    \includegraphics[width=0.46\textwidth]{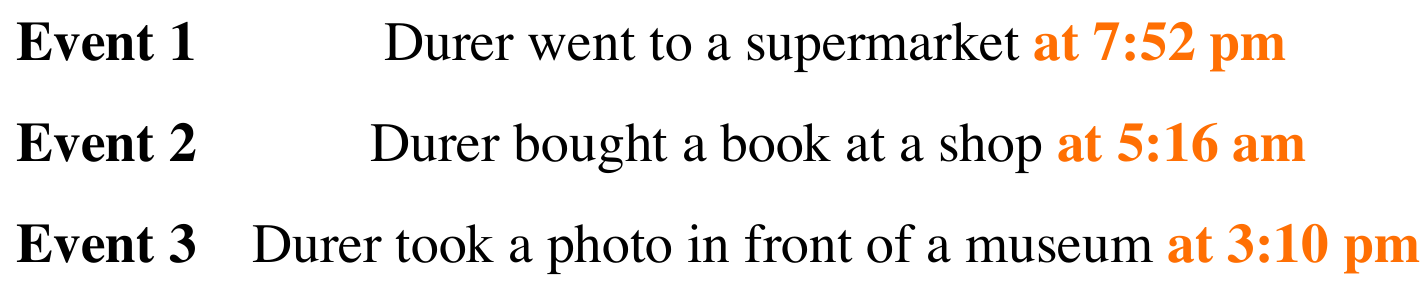}
    \caption{An example of test input event sequences for timex evaluation. The appended timexes in each event, which are 12-hour clock time here, is highlighted.}
    \label{fig:timex_examples}
\end{figure}
\begin{table*}[!t]
    \centering
    \small
    \renewcommand{\tabcolsep}{1.6mm}
    \begin{tabular}{l|cc|cc|cc|cc|cc}
      \toprule
      \multirow{2}{*}{Architecture} & \multicolumn{2}{c|}{Year} & \multicolumn{2}{c|}{Month} & \multicolumn{2}{c|}{Weekday} & \multicolumn{2}{c|}{Hour:Minute (24)} & \multicolumn{2}{c}{Hour:Minute (12)}\\
      & EM & Pairwise  & EM & Pairwise & EM & Pairwise & EM & Pairwise & EM & Pairwise \\
      \midrule
      Random & 26.0 & 53.0 & 21.0 & 51.0 & 18.0 & 52.0 & 18.0 & 50.7 & 13.0 & 52.7 \\
      GPT-2 & 18.0 & 49.0 & 14.0 & 45.3 & 18.0 & 55.3 & 12.0 & 43.3 & 15.0 & 46.7 \\
      GPT-2 Large & 15.0 & 47.3 & 16.0 & 47.7 & 19.0 & 57.7 & 9.0 & 41.7 & 11.0 & 48.7 \\
      \midrule
      TemporalBART & \textbf{93.0} & \textbf{96.7} & 83.0 & 88.7 & 67.0 & 78.0 & \textbf{88.0} & \textbf{93.3} & \textbf{67.0} & \textbf{78.7} \\
      TemporalBART-indexed & 81.0 & 91.3 & \textbf{85.0} & \textbf{90.0} & \textbf{71.0} & \textbf{80.7} & 84.0 & 90.7 & 65.0 & 78.0 \\
      \bottomrule
    \end{tabular}
    \caption{The results of temporal event ordering on the events anchored with various types of timex. The test data used here are length-3 sequences artificially made up with ``die'' events for the ``Year'' timex, and 3 typical daily events as shown in Figure \ref{fig:timex_examples} for other types of timex. The timexes of type ``Year'' are randomly sampled from 1000 to 2100. Our BART-based models significantly outperform the GPT-2 and random baselines, showing that they can capture useful timex-related knowledge.}
    \label{tab:timex_ordering}
\end{table*}
\section{Learning Timex Knowledge}
\label{sec:timex}

The temporal ordering and event infilling tasks correspond to information that we might expect to be encoded by our model pre-training. To test whether our models generalize to slightly more distant temporal phenomena, we examine whether they are able to capture the temporal relationships between timexes. This knowledge has been shown to be hard to learn in temporal relation extraction models \citep{goyal-durrett-2019-embedding}.

\subsection{Evaluation Setup}

The timexes we examine here include years, months, weekdays, 24-hour clock time in ``hour:minute'' format and 12-hour clock time in ``hour:minute am/pm'' format.  We evaluate the ability of our models to order events that are anchored with a timex in their arguments. To prepare the test input event sequences of a given type of timex, we first artificially make up a template event sequence with 3 typical daily events that have no temporal order relations. We then randomly sample 3 different timexes, e.g ``June'', ``May'', ``July'' for ``Month'', and append each of them to the events in the template sequence respectively with proper prepositions. At the end, 100 examples are created with this process for each type of timex. More concrete examples are shown in Figure \ref{fig:timex_examples}. For the baselines, here we use GPT-2 models to do the ordering by using the generation probability to rank all permutations of the input events.

\subsection{Results}
The results are shown in Table \ref{tab:timex_ordering}. First, we examine the results of the GPT-2 models. In general both the unsupervised GPT-2 (the medium model) and GPT-2 large perform worse than the random baseline, indicating that they have a limited ability to order timexes. Our BART-based models achieve stronger results. The results are strongest on years. 
For 12-hour clock time, even though the model has to make a challenging link between the temporal knowledge on ``am' and ``pm'' and numerical comparisons, both of the BART-based models still performs significantly better the random baseline.

\begin{table*}[t]
    \centering
    \small
    \renewcommand{\tabcolsep}{2.5mm}
    \renewcommand{\arraystretch}{1.5}
    \begin{tabular}{l|p{4.5cm}|p{4.5cm}|p{4.5cm}}
      \toprule
      \multirow{2}{*}{\#1} & \multicolumn{3}{c}{\textbf{Scenario}: \texttt{[INSERTED EVENT]} he tell me Happy Birthday} \\
      \cmidrule{2-4}
      & \textbf{GPT-2}: A New England University study reveals the effect of the American tax code on college tuition . & \textbf{Infilling GPT-2}: He then turned to me and his mom and said, `` Good morning . & \textbf{TemporalBART}: He looked at me in the most amazing of ways, as if he were wishing me a Happy Birthday . \\
      \toprule
      \multirow{2}{*}{\#2} & \multicolumn{3}{c}{\textbf{Scenario}: \texttt{[INSERTED EVENT]} Then he bought it with the card .} \\
      \cmidrule{2-4}
      & \textbf{GPT-2}: A federal judge dismissed a lawsuit Wednesday from the family of an elderly man who died after inhaling a toxic gas during an air show at the Lincoln Memorial . & \textbf{Infilling GPT-2}: He put the card in his wallet . & \textbf{TemporalBART}: He pulled out a small gift card from his wallet . \\
      \toprule
      \multirow{2}{*}{\#3} & \multicolumn{3}{p{14.5cm}}{\textbf{Scenario}: Kim hunkered down at home as a hurricane blew through town . \texttt{[INSERTED EVENT]} She went to sleep hoping all would be well . She hoping all would be well .} \\
      \cmidrule{2-4}
      & \textbf{GPT-2}: It wasn't until his brother and mother returned from vacation that he found out that a hurricane had struck, bringing the death toll from Hurricane Andrew in 2012 up to 24 and leaving many people without electricity for weeks . & \textbf{Infilling GPT-2}: Kim slept soundly . & \textbf{TemporalBART}: Kim turned the TV to catch the latest news . \\
      \toprule
      \multirow{2}{*}{\#4} & \multicolumn{3}{p{14.5cm}}{\textbf{Scenario}: Tony needed to buy his grandma a birthday present . He went to her favorite bakery . The owner told Tony how to make it himself . \texttt{[INSERTED EVENT]} His elated grandma couldn't tell the difference .} \\
      \cmidrule{2-4}
      & \textbf{GPT-2}: Tony got mad and left the bakery . & \textbf{Infilling GPT-2}: Tony went to the grocery store . & \textbf{TemporalBART}: make a loaf that looked like grandma's . \\
      \toprule
      \multirow{2}{*}{\#5} & \multicolumn{3}{p{14.5cm}}{\textbf{Scenario}: \texttt{[INSERTED EVENT]} He decided to use the batteries in his fire detector . He use the batteries in his fire detector .} \\
      \cmidrule{2-4}
      & \textbf{GPT-2}: I'm an independent developer who's worked for both big and small companies . & \textbf{Infilling GPT-2}: He find a place to charge the fire detector batteries . & \textbf{TemporalBART}: He see the batteries in his alarm clock were dead . \\
      \bottomrule
    \end{tabular}
    \caption{More examples of the infilled events generated by GPT-2, infilling GPT-2 and TemporalBART respectively. Scenarios are the temporally-ordered input events fed into the models, with the events separated by periods, and the insertion position specified by \texttt{[INSERTED EVENT]} in this figure. The second row in each example shows the infilled event generated by each model.}
    \label{tab:more_infilling_examples}
\end{table*}
\section{Examples for Event Infilling}
In Table \ref{tab:more_infilling_examples}, we demonstrate more examples of the infilled events generated by GPT-2, infilling GPT-2 and TemporalBART given the seed event sequences from CaTeRS. In general, while the events output by TemporalBART are coherent and temporally-sensible, those from the GPT-2 models has a worse quality in terms of the temporality. Note that the nature of the event representation does not necessarily guarantee a grammatical sentence when the event is rendered in surface order.

\end{document}